\documentclass{article}

\usepackage{microtype}
\usepackage{graphicx}
\usepackage{subfigure}
\usepackage{booktabs}
\usepackage{hyperref}
\usepackage{amsmath,amsfonts}
\DeclareMathOperator*{\argmin}{arg\,min}

\usepackage[dynnworkshop]{icml2022}

\icmltitlerunning{Dynamic Split Computing for Efficient Deep Edge Intelligence}

\begin{document}

\twocolumn[
\icmltitle{Dynamic Split Computing for Efficient Deep Edge Intelligence}

\begin{icmlauthorlist}
\icmlauthor{Arian Bakhtiarnia}{aarhus}
\icmlauthor{Nemanja Milošević}{novisad2}
\icmlauthor{Qi Zhang}{aarhus}
\icmlauthor{Dragana Bajović}{novisad1}
\icmlauthor{Alexandros Iosifidis}{aarhus}
\end{icmlauthorlist}

\icmlaffiliation{aarhus}{DIGIT, Department of Electrical and Computer Engineering, Aarhus University, Denmark}
\icmlaffiliation{novisad1}{Faculty of Technical Sciences, University of Novi Sad, Serbia}
\icmlaffiliation{novisad2}{Faculty of Sciences, University of Novi Sad, Serbia}

\icmlcorrespondingauthor{Arian Bakhtiarnia}{arianbakh@ece.au.dk}

\icmlkeywords{Split Computing, Collaborative Intelligence, Edge Intelligence, Computer Vision}

\vskip 0.3in
]

\printAffiliationsAndNotice{}

\begin{abstract}
Deploying deep neural networks (DNNs) on IoT and mobile devices is a challenging task due to their limited computational resources. Thus, demanding tasks are often entirely offloaded to edge servers which can accelerate inference, however, it also causes communication cost and evokes privacy concerns. In addition, this approach leaves the computational capacity of end devices unused. Split computing is a paradigm where a DNN is split into two sections; the first section is executed on the end device, and the output is transmitted to the edge server where the final section is executed. Here, we introduce dynamic split computing, where the optimal split location is dynamically selected based on the state of the communication channel. By using natural bottlenecks that already exist in modern DNN architectures, dynamic split computing avoids retraining and hyperparameter optimization, and does not have any negative impact on the final accuracy of DNNs. Through extensive experiments, we show that dynamic split computing achieves faster inference in edge computing environments where the data rate and server load vary over time.
\end{abstract}

\section{Introduction}
\label{introduction}

\begin{figure*}
\begin{center}
\begin{tabular}{ c c c c }
\includegraphics[width=0.22\textwidth]{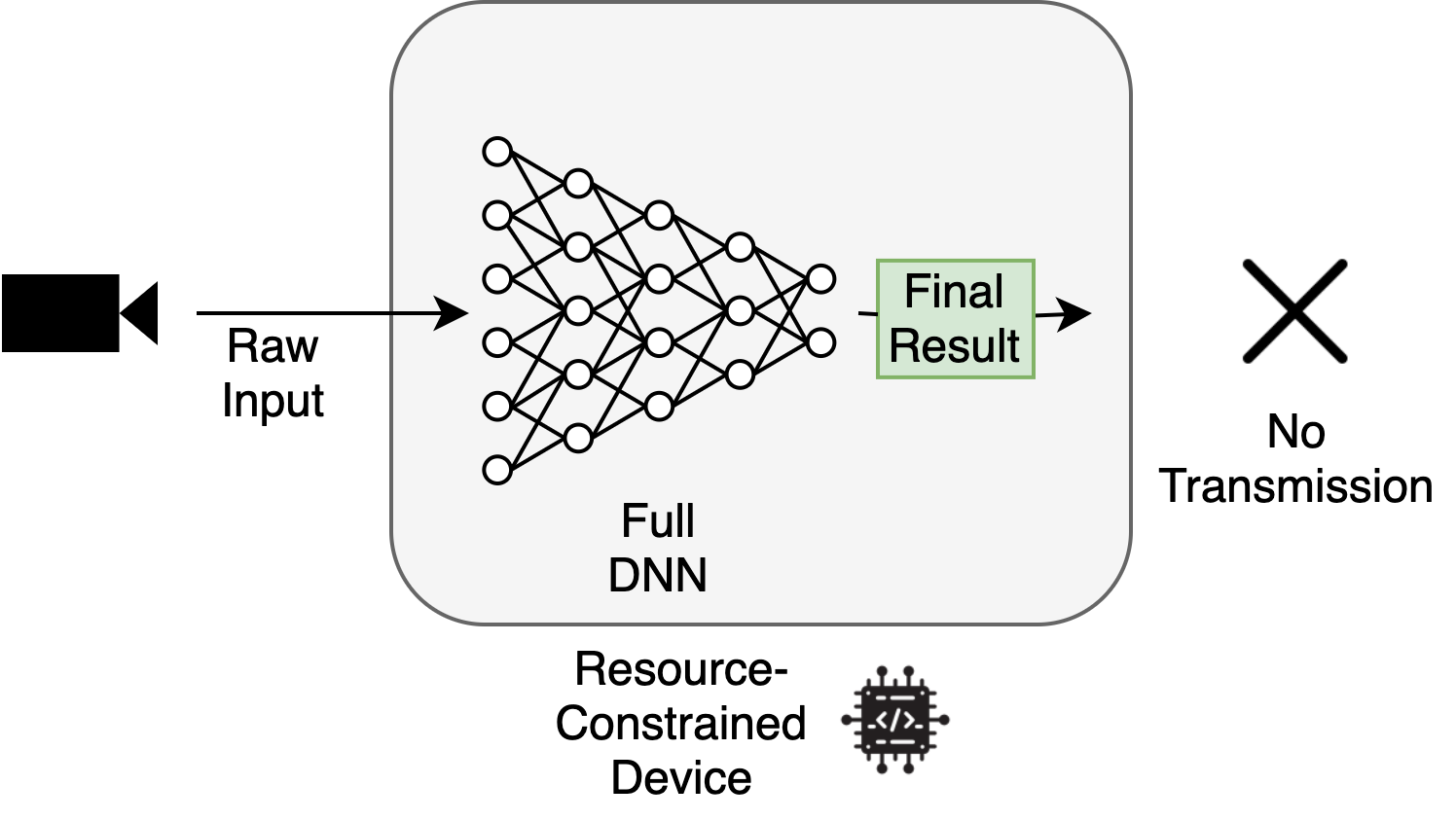} & \includegraphics[width=0.22\textwidth]{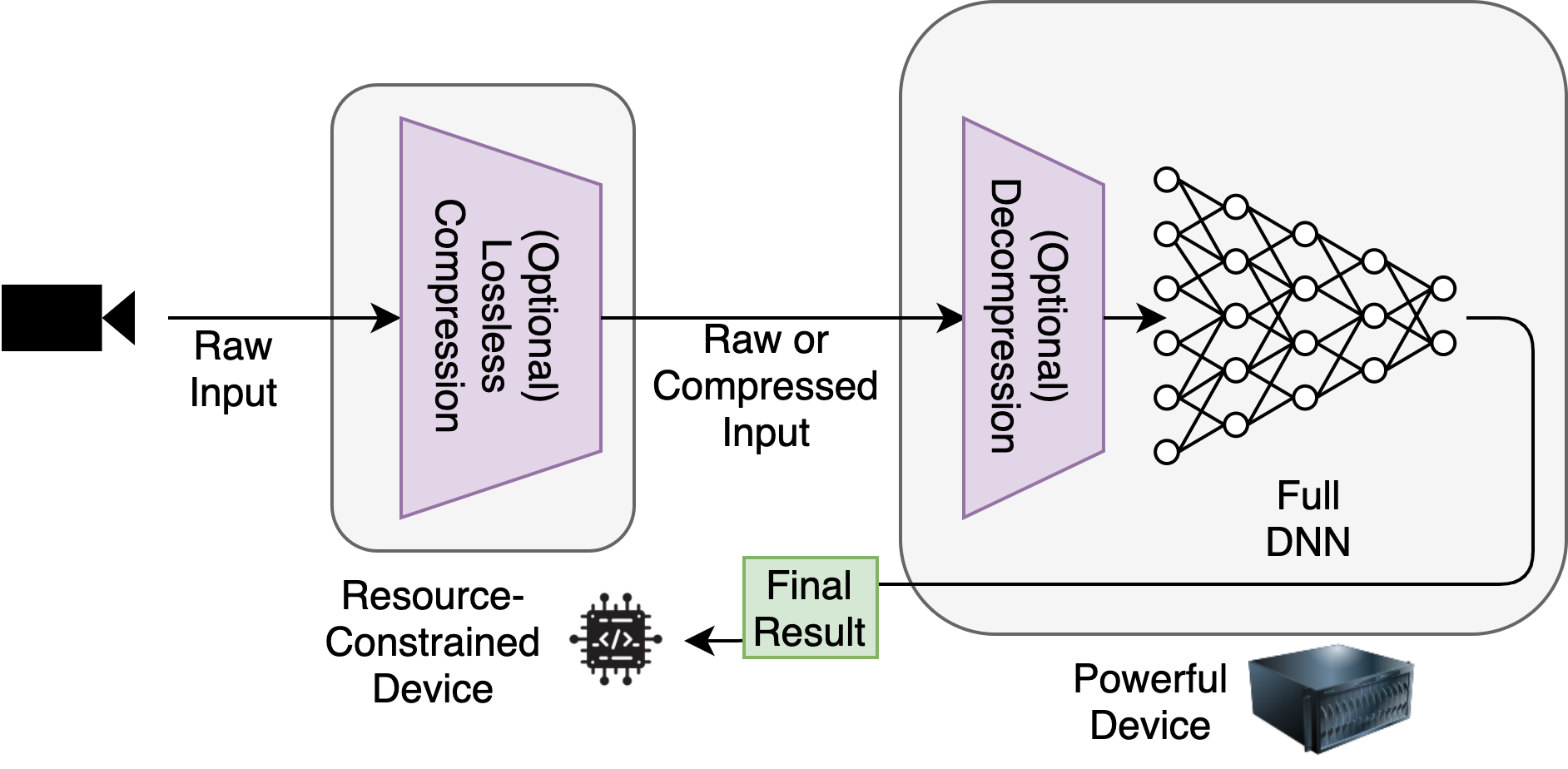} & \includegraphics[width=0.22\textwidth]{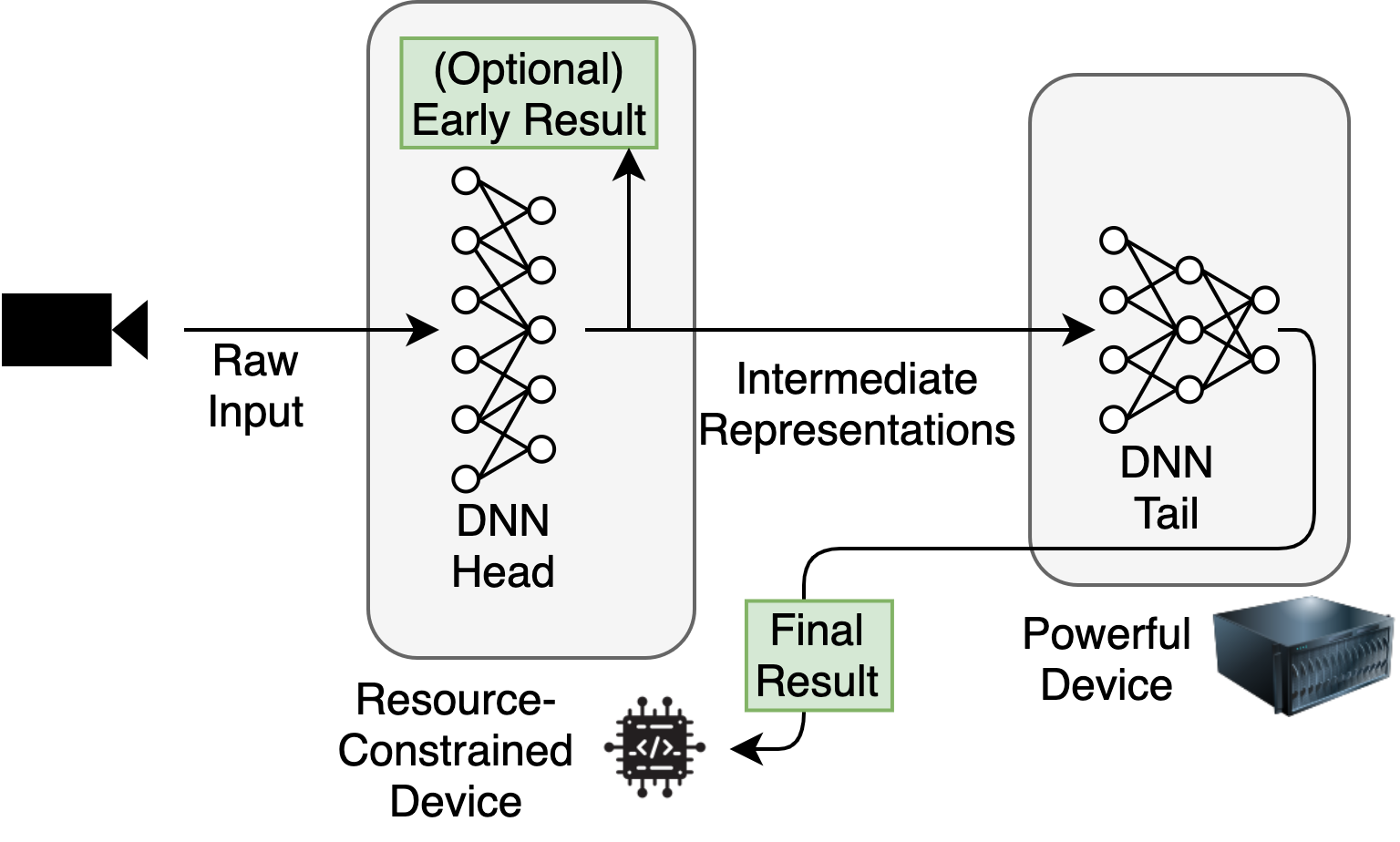} & \includegraphics[width=0.22\textwidth]{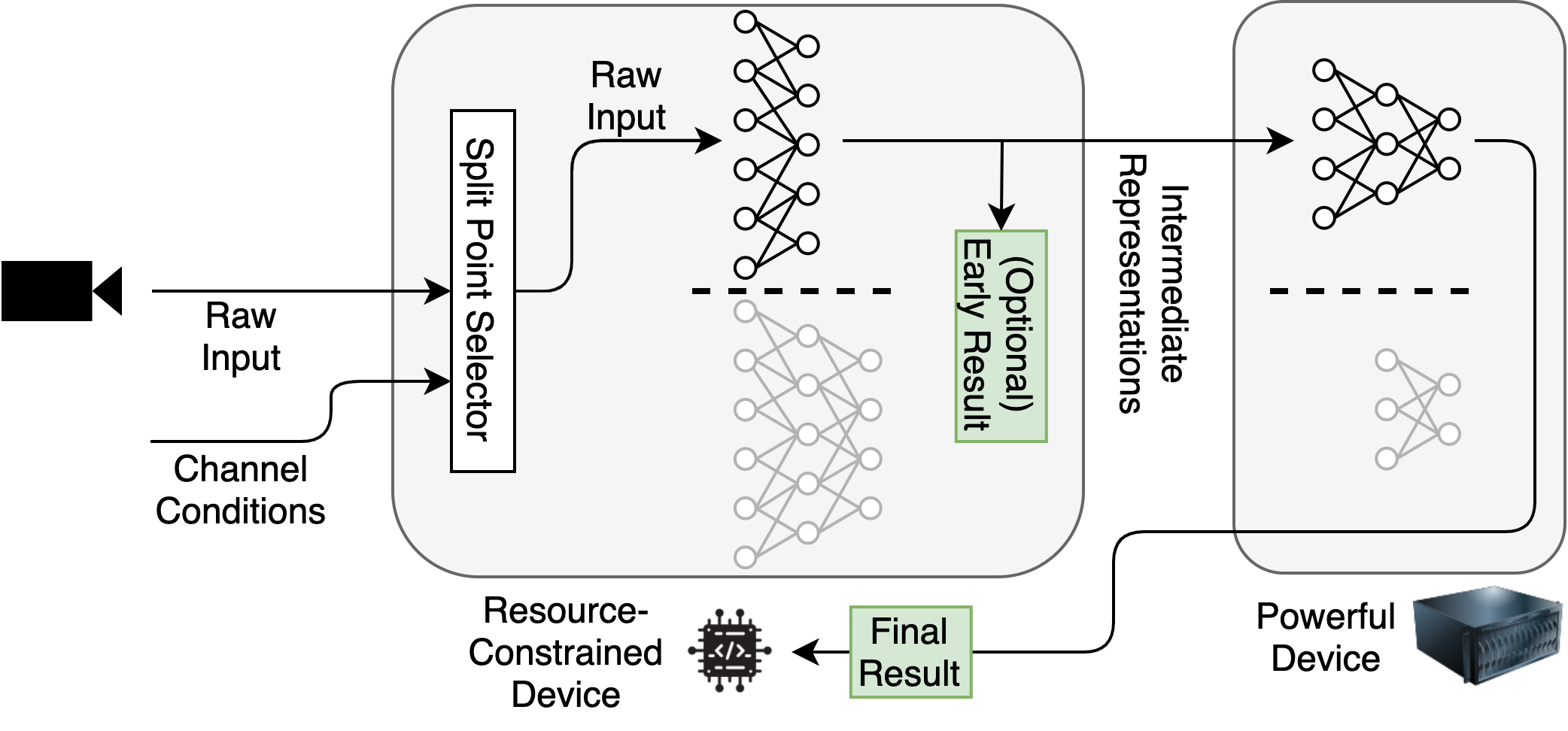}\\
(a) & (b) & (c) & (d)\\
\end{tabular}
\end{center}
\caption{Overview of (a) no-offloading; (b) full-offloading; (c) split computing; and (d) dynamic split computing approaches.}
\label{fig:overview}
\end{figure*}

The combination of deep learning and Internet of Things (IoT) has tremendous applications in fields such as healthcare, smart homes, transportation and industry \cite{8932460}. However, deep learning models typically contain millions or even billions of parameters, making it difficult to deploy these models on resource-constrained devices. One solution is to offload the computation to an edge or cloud server \cite{8976180}, as shown in Figure \ref{fig:overview} (b). However, since the size of the inputs to deep learning models can be massive, particularly images and videos, this approach consumes a lot of bandwidth and energy, and leads to delays. Moreover, even though IoT devices are limited, they still possess computational capabilities that remain unused when the entire computation is offloaded, and utilizing these capabilities would reduce the load on the servers. In addition, in applications that process personal data such as health records, or in audio or visual streams with voice activity or human presence, privacy regulations such as European Union's GDPR~\cite{GDPR2018} or United States' HIPAA~\cite{hipaa96} may apply. These regulations typically forbid direct access to non-anonymized data, leaving the options to either anonymize the data at the cost of additional computation and higher latency, or process the data at the source.

\textit{Split computing}, depicted in Fig. \ref{fig:overview} (c), alleviates these issues by splitting the deep model into a \textit{head} section and a \textit{tail} section \cite{2103.04505}. The head model is executed on the device, and its output (the intermediate representation at that particular layer of the deep network) is transmitted to the server, then processed by the tail model to obtain the final output. In a way, split computing is a \textit{partial offloading} of the computation, as opposed to the \textit{full-offloading} approach. Another benefit of split computing over full-offloading is that it can be used as a privacy preserving technique since intermediate representations are being transmitted instead of the actual inputs, and the original inputs cannot be easily reconstructed from the intermediate representations \cite{8416417}. In addition, split computing can be combined with early exiting in order to obtain an early result on the device \cite{Scardapane2020, 2103.04505}, as illustrated in Figure \ref{fig:overview} (c), which is useful when transmission takes longer than expected.

Since split computing aims to decrease the communication cost, \textit{natural bottlenecks}, which are the layers of the deep network where the size of the intermediate representation is smaller than the input size, can be used as splitting points for deep learning models. In this paper, we show that unlike older popular models, state of the art models such as EfficientNet \cite{pmlr-v97-tan19a, 2104.00298} possess many natural bottlenecks. Based on this fact, we propose a method called \textit{dynamic split computing} where the best splitting point is automatically and dynamically determined based on input and channel conditions, as shown in Figure \ref{fig:overview} (d). Since the underlying deep learning model is not modified, dynamic split computing can be used as a plug-and-play method, meaning it can be employed without domain knowledge about the particular deep learning models that are being used. It is important to note that dynamic split computing is a complimentary efficient inference method that can be used in combination with other approaches, including model compression techniques such as pruning and quantization \cite{Choudhary2020}, as well as dynamic inference methods such as early exiting \cite{2106.15183}.\footnote{Our code is available at \url{https://gitlab.au.dk/maleci/dynamicsplitcomputing}.}

\section{Related Work}\label{S:RelatedWork}

Several approaches for speeding up the inference of deep neural networks (DNNs) on resource-constrained devices exist in the literature. \textit{Local computing} performs the entire computation on the device, yet modifies the architecture of the neural network in order to decrease the required computation, while causing a minimal negative impact on the accuracy. \textit{Lightweight models} such as MobileNet \cite{1704.04861, Sandler_2018_CVPR, Howard_2019_ICCV} are specifically designed to be deployed on such limited devices, whereas \textit{model compression} techniques \cite{8253600} alter existing architectures in order to make them more lightweight, for instance, \textit{pruning} removes the less impactful parameters (weights) of the neural network; \textit{quantization} uses less bits to represent each parameter \cite{LIANG2021370}; and \textit{knowledge distillation} aims to train a more compact model to reproduce outputs similar to a given larger neural network \cite{Gou2021}.

\textit{Dynamic inference} methods \cite{2102.04906} can alter the architecture of existing neural networks to adapt their inference time at the cost of accuracy, meaning they will produce more accurate outputs the longer they are allowed to execute. Various approaches to dynamic inference exist, such as \textit{early exiting} \cite{Scardapane2020}, where early exit branches are added after intermediate layers of a DNN that produce an output similar to the final output; \textit{layer skipping} \cite{1603.08983, 2107.05407, Wang_2018_ECCV}, where the execution of some of the DNN layers are skipped; and \textit{channel skipping} \cite{DBLP:conf/iclr/GaoZDMX19}, where less impactful channels of convolutional layers are ignored.

Even with local computing techniques, many high-performing DNNs exceed the computational capacity of devices, especially when the output is expected within a strict deadline. In such cases, the computation can be offloaded to external servers. When the computation of a DNN is offloaded, the inputs must be transmitted from a device to a server, yet this can introduce massive delays during data transmission, particularly when the input size is large, which may defeat the original purpose of speeding up the computation. This has led to a recent emerging paradigm called \textit{edge computing} \cite{8030322} where the computation is offloaded to \textit{edge servers} located much closer to end devices compared to cloud servers which are often located in data centers. Even though edge computing reduces the transmission delay, it still has some drawbacks. First, since the original inputs are being transmitted over a network, privacy issues arise. Furthermore, since typically multiple end devices are connected to the same edge server, if all of them offload their computation simultaneously, the edge server may experience a high load while the computational resources of each end device remain unused.

\textit{Split computing} \cite{2103.04505} (also known as \textit{collaborative intelligence}) is an alternative approach that provides a balance between local computing and full-offloading, where some layers of the DNN are executed on the end device and the intermediate output is then sent over to the edge server where it is processed by the rest of the DNN layers. When the splitting point is chosen such that the size of the intermediate representation is lower than the input size, the transmission delay will consequently be lower than that of full-offloading.

However, not all deep learning models possess such natural bottlenecks, and even if they exist, they may be located in the final layers of the network where the bulk of the computation has already been carried out, and therefore it would not be sensible to offload the remaining computation. For instance, widely used models such as ResNet \cite{He_2016_CVPR} and Inception \cite{Szegedy_2016_CVPR} do not contain natural bottlenecks in their early layers \cite{2103.04505}. In such cases, \textit{bottleneck injection} can be used, where the architecture of the network is modified to artificially insert a bottleneck \cite{2103.04505}. However, this approach requires time-consuming operations such as retraining the model and optimizing hyperparameters such as the size of the inserted bottleneck. Furthermore, there is no guarantee that the new architecture can obtain an accuracy comparable to that of the original architecture, particularly when a limitation such as a small bottleneck is imposed. Therefore, bottleneck injection is far from ideal.

\section{Dynamic Split Computing}\label{S:Method}

We assume a trained high-performing DNN is to be deployed on a device with access to a server, where the data rate of the communication channel and the number of inputs in the batch may vary. The variations in the data rate may be due to fluctuations in wireless channel or traffic congestion, and the variations in batch size may occur due to different workloads at different times. The goal of our method is to optimize the end-to-end inference time by dynamically detecting the best splitting point for a given DNN based on the communication channel state and batch size. Since we aim to design our method in a ``plug-and-play'' manner, such that it can be deployed in edge computing systems without creating trade-offs involving the accuracy or the hassle of retraining, we avoid altering the underlying architecture or lossy compression techniques that may affect the accuracy.

Formally, neural networks can be formulated as $ f(x) = f_L(f_{L-1}( \dots f_1(x))) $ where $ x $ is the input, $ L $ is the total number of layers in the neural network and $ f_i $ is the operation performed at layer $ i $. The intermediate representation at layer $ i $, which is the output of the $ i $-th layer is recursively formulated as $ h_i = f_i(h_{i-1}) $ where $ h_0 = x $ is the input. Based on this notation, with split computing at layer $ j $, the head and tail parts of the DNN are denoted by $ f^h(x) = f_j(f_{j-1}( \dots f_1(x))) $ and $ f^t(h_j) = f_L(f_{L-1}( \dots f_{j+1}(h_j))) $, respectively, and $ h_j $ is the intermediate representation that is transmitted.

The first step is to find the natural bottlenecks of the DNN by calculating the compression ratio $ c_l = |h_l| / |x| $ for each layer $ l $ where $ |h_l| $ and $ |x| $ denote the size of intermediate representation at layer $ l $ and the input size, respectively. If $ c_l < 1 $, layer $ l $ is a natural bottleneck of the DNN. However, not all natural bottlenecks are useful in split computing. We define $ T^h_{i,j} $ and $ T^t_{i,j} $ as the inference time from layer $ i $ up to and including layer $ j $ ($ i < j $) of the deep neural network measured on the device and the server, respectively. When layers $ m $ and $ n $ ($ m < n $) have the same compression ratio, in other words when $ c_m = c_n $, the total end-to-end inference time with split computing at layer $ m $ and layer $ n $ are
\begin{eqnarray}
T_m &=& T^h_{1, m} + c_m T_{\text{FO}} + T^t_{m + 1, n} + T^t_{n + 1, L},
\label{eq:two_times_m} \\
T_n &=& T^h_{1, m} + T^h_{m + 1, n} + c_n T_{\text{FO}} + T^t_{n + 1, L}.
\label{eq:two_times_n}
\end{eqnarray}
where $ T_{\text{FO}} $ is the transmission time of the entire input in full-offloading. Assuming the computational resources of the server are greater than that of the device, then $ T^h_{m + 1, n} > T^t_{m + 1, n} $, thus it is favorable to choose the earlier layer as splitting point. 
Consequently, only natural bottlenecks with compression ratio lower than all previous natural bottlenecks are useful. We call such bottlenecks \textit{compressive}. Compressive natural bottlenecks are defined by
\begin{equation}
C = \{j | c_j < 1, c_j < c_i \; \forall i < j\}.
\label{eq:compressive}
\end{equation}
The total end-to-end inference time for a given batch of inputs when the splitting point of the network is $l$ is
\begin{equation}
T_l = T^h_{1, l} + \frac{D \medspace c_l}{r} + T^t_{l + 1, L},
\label{eq:total_time}
\end{equation}
where $ D $ is the data size of the original input, $ c_l $ is the compression ratio of the intermediate representation at layer $ l $ and $ r $ is the data rate of the communication channel. When inputs are images or video frames, the total load in bytes can be calculated as $D = B W H C$, where $ B $ is the batch size, $ W $ and $ H $ are the width and height of the images, and $ C $ is the number of channels in the images, for instance, $ C = 3 $ for color images and $ C = 1 $ for grayscale.

We define the end-to-end inference time in case of no-offloading as $T_L = T^h_{1, L}$ and in case of full-offloading as
\begin{equation}
T_0 = \frac{D}{r} + T^t_{1, L}.
\label{eq:full_offloading_time}
\end{equation}
Therefore, the optimal splitting point $ s_{opt} $ can be determined by optimizing for
\begin{equation}
s_{opt} = \argmin_{l \in \{0 \dots L\}}(T_l).
\label{eq:best_split}
\end{equation}
Dynamic split computing finds the optimal split location for a given data rate and batch size based on Eq. \eqref{eq:best_split} at each time step, and switches to that configuration. When full-offloading cannot be used, for instance, due to privacy requirements, the range in Eq. \eqref{eq:best_split} is reduced to $ \{0 \dots L-1\} $. Note that based on previous arguments, only compressive natural bottlenecks need to be investigated, therefore once all compressive natural bottlenecks are identified, we calculate the optimal splitting point for each batch size and data rate by measuring the inference time of head and tail models for each compressive bottleneck. It is important to note that the relationship between inference time of head or tail model and batch size is not strictly linear, therefore it needs to be measured for each batch size. Additionally, when the data rate is too low, it may not be sensible to use any form of offloading since it introduces too much delay. Therefore, dynamic split computing considers the no-offloading option alongside the optimal splitting point and switches between split computing and no-offloading when necessary.

Since different applications and environments may have different ranges for data rate and batch size and a unique pattern for their variations, we need a method to measure how beneficial dynamic split computing is in each specific case. We define a scenario as a sequence of the state of the environment throughout time, i.e., $S = ((B_1, r_1), (B_2, r_2) \dots, (B_T, r_N))$, where $ B_i $ and $ r_i $ are the batch size and data rate at time step $ i $, respectively. The relative average gain of dynamic split computing in terms of end-to-end inference time over a specific method, for instance, static split computing at a specific location, can then be calculated by
\begin{equation}
G = \frac{1}{N} \sum_{1 \leq i \leq N}\frac{|T_{s_{opt}}(B_i, r_i) - T^{\text{SS}}(B_i, r_i)|}{T^{\text{SS}}(B_i, r_i)},
\label{eq:gain}
\end{equation}
where $ T_{s_{opt}}(B_i, r_i) $ and $ T^{\text{SS}}(B_i, r_i) $ are the end-to-end inference time using dynamic and static split computing, respectively, with batch size $ B_i $ and data rate $ r_i $.

\section{Results}\label{S:Results}

We investigate 14 modern DNN architectures: seven variations of EfficientNetV2 \cite{2104.00298} and seven variations of EfficientNetV1 \cite{pmlr-v97-tan19a}. All these architectures were originally designed for image classification and have since been applied to various other problems such as speech recognition \cite{10.1145/3404716.3404717}. The accuracy of these architectures on the ImageNet dataset \cite{5206848} ranges from 77.1\% to 85.7\%.

First, we find the compressive natural bottlenecks for each architecture. The number of natural bottlenecks in these architectures ranges from 15 to 68, three to four of which are compressive. For comparison, VGG-16 \cite{1409.1556}, which is an older architecture, has only 5 natural bottlenecks. Subsequently, we find the optimal splitting point for each architecture in a wide range of states. We check data rates ranging from 1 MBps to 128 MBps and batch sizes of 1 to 64. Some larger models such as EfficientNetV2-L run into memory issues with large batch sizes, therefore, we reduce the maximum batch size to 32 or 24 in such cases. For the edge server, we use an Nvidia 2080 Ti GPU, and in order to simulate a resource-constrained device, we underclock the same type of GPU to 300 MHz (the normal GPU frequency is around 1800 MHz).

\begin{figure}[htbp]
\centerline{\includegraphics[width=0.48\textwidth]{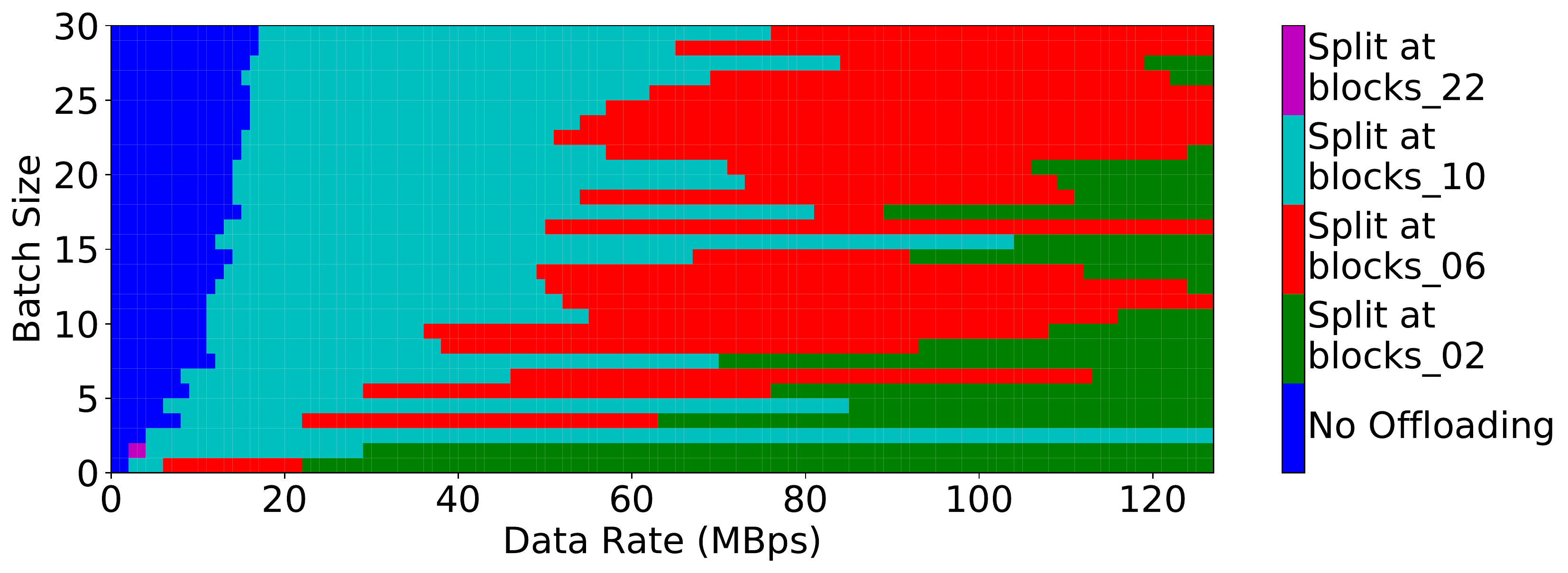}}
\caption{Optimal split location based on batch size and data rate for the EfficientNetV1-B4 architecture.}
\label{efficientnet-b4-gpu}
\end{figure}

\begin{figure}[htbp]
\centerline{\includegraphics[width=0.48\textwidth]{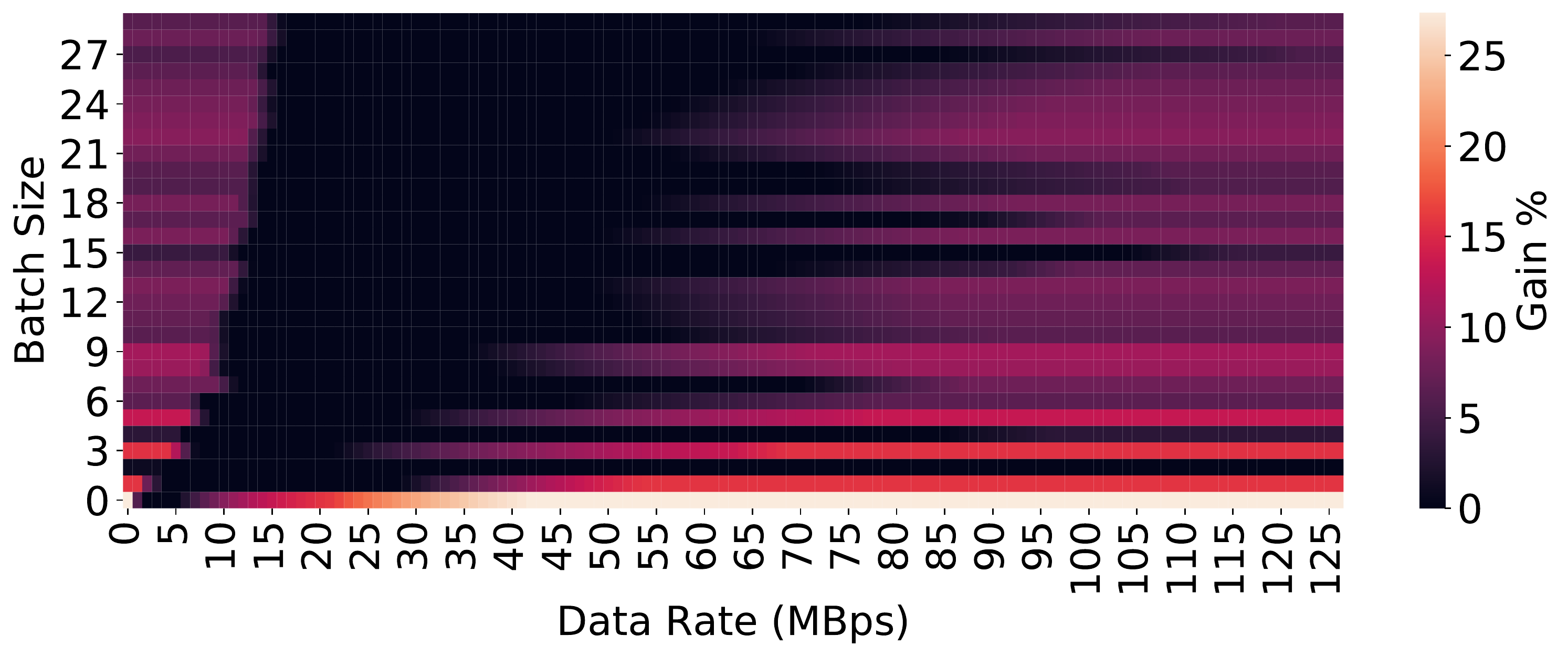}}
\caption{The relative gain of dynamic split computing in terms of end-to-end inference time over static split computing at block 10 in the EfficientNetV1-B4 architecture.}
\label{gain-block-10}
\end{figure}

The results for the EfficientNetV1-B4 architecture are shown in Fig. \ref{efficientnet-b4-gpu}, where for each state (data rate and batch size), the optimal split location derived based on Equation \ref{eq:best_split} is specified. It can be observed that each compressive bottleneck is an optimal split location in several states. Moreover, no-offloading is the optimal solution in some other states. Therefore, dynamically switching between split locations (as well as no-offloading) based on the state of the communication channel improves inference speed. This is also the case with the other 13 investigated architectures. The relative gain of dynamic split computing over split computing at a fixed location (block 10) for the EfficientNetV1-B4 architecture in terms of inference speed is shown in Fig. \ref{gain-block-10}. This figure can be used to derive the gain of dynamic split computing compared to another method for a specific scenario based on on Equation \ref{eq:gain}. Notice that in states where split computing at block 10 is optimal, dynamic split computing swiches to this method and thus has no advantage over it, whereas dynamic split computing obtains some gain everywhere else by switching to a different method.

\section{Conclusion}
In this paper, we showed that dynamic split computing improves in inference time over both no-offloading and split computing with a fixed split location. Moreover, as opposed to full-offloading, dynamic split computing can decrease the load on the server by performing parts of the computation on the device. Finally, by transmitting intermediate representations instead of inputs, dynamic split computing circumvents privacy issues that arise when using full-offloading.

\section*{Acknowledgements}
The work received funding by the European Union’s Horizon 2020 research and innovation programme under grant agreement No 957337, and by the Danish Council for Independent Research under Grant No. 9131-00119B. 

\bibliography{main}
\bibliographystyle{icml2021}

\end{document}